\documentclass[10pt]{extarticle}
\usepackage[utf8]{inputenc}
\usepackage{geometry}
\usepackage{graphics}
\usepackage{amsfonts}
\usepackage{indentfirst}
\usepackage{hyperref}
\usepackage{amssymb}
\usepackage{color,xcolor}
\usepackage{graphicx,epstopdf}
\usepackage{epsfig}
\usepackage{subcaption}
\usepackage{mathrsfs}
\usepackage{bbm}
\usepackage{chngpage}
\usepackage{diagbox}
\usepackage{multirow}
\usepackage{multicol}
\usepackage{dsfont}
\usepackage{bm}
\usepackage{amsmath}
\usepackage{amssymb,amsmath,amsthm,mathrsfs}%
\usepackage{exscale}
\usepackage{tikz-cd}
\usepackage{relsize}
\usepackage{amssymb}
\usepackage{multirow}
\usepackage{booktabs}
\usepackage{flushend}

\hypersetup{colorlinks=true,linkcolor=blue,anchorcolor=green,citecolor=red}

\usepackage[
    natbib=true,
    style=numeric,
    sorting=none
]{biblatex}

\theoremstyle{definition}%{remark}{\plain}

\newtheorem{definition}{Definition}[section]

\newtheorem{corollary}{Corollary}[section]

\newcommand{\mi}{\mathrm{i}} %% roman "i"

\usepackage[font=scriptsize,labelfont=bf]{caption}

\bibliography{./references.bib}

\newgeometry{vmargin={15mm}, hmargin={15mm,15mm}}   %

\title{A Hybrid Complex-valued Neural Network Framework\\ with Applications to Electroencephalogram (EEG)}
\usepackage{leading}
\leading{11pt}
\begin{document}

\author{Hang Du\thanks{Hang Du and Xiaogang Wang are with the Department of Mathematics and Statistics, York University, Toronto, M3J 1L1, Canada (email: beretta@yorku.ca, stevenw@yorku.ca).}, Rebecca Pillai Riddell\thanks{Rebecca Pillai Riddell is with the Department of Psychology, York University, Toronto, M3J 1L1, Canada (email: rpr@yorku.ca).}, Xiaogang Wang\footnotemark[1]}

\maketitle

\begin{abstract}
In this article, we present a new EEG signal classification framework by integrating the complex-valued and real-valued Convolutional Neural Network(CNN) with discrete Fourier transform (DFT). The proposed neural network architecture consists of one complex-valued convolutional layer, two real-valued convolutional layers, and three fully connected layers. Our method can efficiently utilize the phase information contained in the DFT. We validate our approach using two simulated EEG signals and a benchmark data set and compare it with two widely used frameworks. Our method drastically reduces the number of parameters used and improves accuracy when compared with the existing methods in classifying benchmark data sets, and significantly improves performance in classifying simulated EEG signals.

\end{abstract}

\section{Introduction}

EEGs are used in many health care institutions to monitor and record electrical activity in the brain using electrodes placed on the scalp. Analysis of EEG recordings is a crucial first step to making a clinical diagnosis of epilepsy, severity of anesthesia, coma, encephalopathy, and brain death~\cite{yamada_practical_2017}. Traditionally, clinicians analyze EEG signals by visual inspection, which is usually time-consuming and can be affected easily by clinicians' subjectivity~\cite{main_compare}. Thus automatic detection systems based on machine learning algorithms have been developed to evaluate and classify EEG signals~\cite{RAMGOPAL_seizure_2014}.

Due to the specific properties of EEG signals, such a classification task can be more broadly described as non-stationary signal classification. Traditionally, in non-stationary signal classification, the first step is to extract the useful features and then use classifiers like Support Vector Machine(SVM), K-Nearest Neighbors(KNN), Regression, and so on to differentiate classes \cite{non-stationary1998, ns_svm_2002, tf_svm_2014, GLCM_2016}.   Many techniques have been proposed for EEG signal classification. Hassanpour \textit{et~al.} \cite{hassanpour_timefrequency_2004} use modified B-distribution to characterize low-frequency EEG patterns and apply singular value decomposition (SVD) on the time-frequency distribution matrix to detect seizures in newborns and obtain encouraging results. In \cite{tfa_dnn}, Tzallas \textit{et~al.} extract features using Cohen's class Time-Frequency Representation (TFR) and power spectrum density, then use Artificial Neural Networks as the classifier to identify epileptic seizures in three benchmark EEG datasets. Boashas \textit{et~al.} \cite{QTFD_2015} use quadratic time-frequency distributions (TFDs) with extended features and matrix decomposition with SVM as the binary classifier to detect newborn EEG abnormalities. In \cite{motion_eeg}, the authors use wavelet energy and wavelet entropy as features and KNN as the classifier. In \cite{o.k._time-domain_2019}, the authors propose exponential energy as a new feature, and combine it with other commonly used entropy and energy features, then use SVM to classify epileptic EEG signals.

\begin{figure*}[htp]
     \centering
     %%%%%%%%%%%%%%%%%%%%%%%%%%%%%%%%%%%%%%%%%
     \begin{subfigure}[t]{0.6\textwidth}
         \centering
         \includegraphics[width=\textwidth]{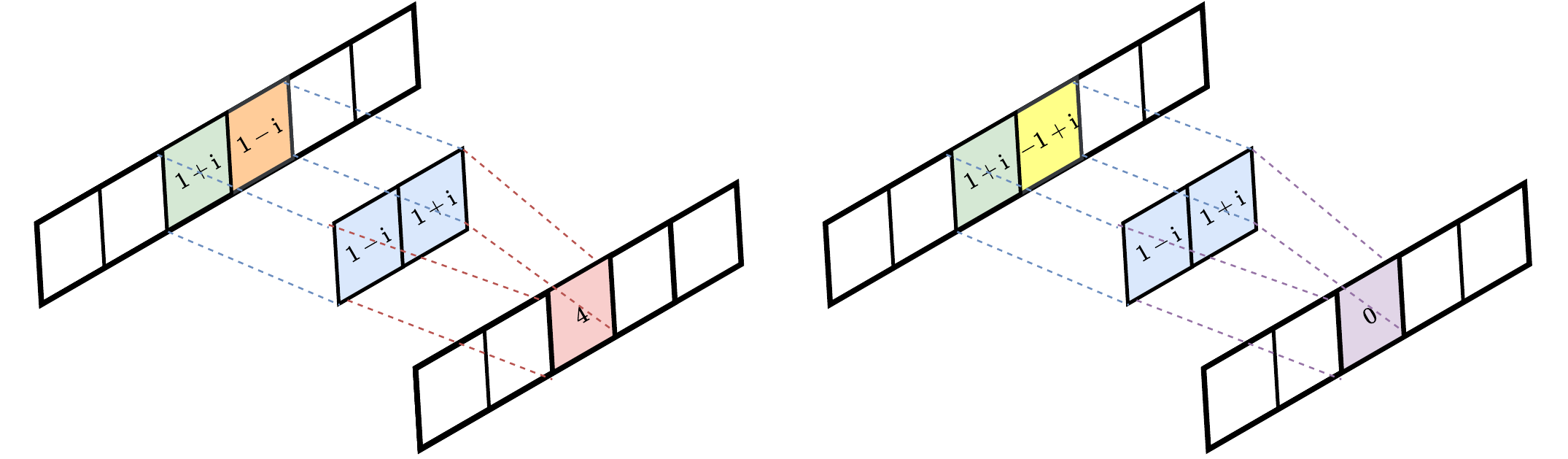}
         \caption{}
          \label{aaaa}
     \end{subfigure}
    %  \hfill
     \tikz{\draw[-,black, densely dashed](0,-1.05) -- (0,2.4);}
     \begin{subfigure}[t]{0.3\textwidth}
         \centering
         \includegraphics[width=\textwidth]{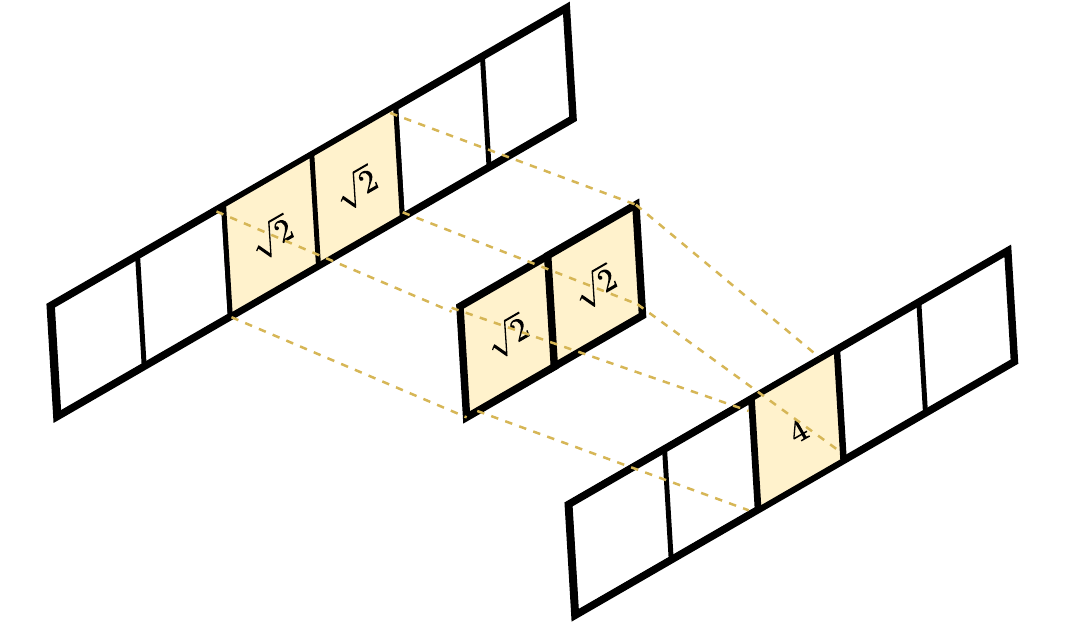}
         \caption{}
          \label{bbbb}
     \end{subfigure}
     %%%%%%%%%%%%%%%%%%%%%%%%%%%%%%%%%%%%%%%%%%%%%%%%%
    %  \vspace{-0.1in}
        \caption{The convolution operation (a) on complex numbers, (b) on modulus (amplitude) only.}
        \label{real_complex_conv_compare}
        \vspace{-0.1in}
\end{figure*}

In the recent decade, it has been shown that methods based on deep learning can yield better performance than conventional methods \cite{main_compare,tf_spatial_conv,Gao_Gao_Chen_Liu_Zhang_2020,Gao_Dang_Wang_Hong_Hou_Ma_Perc_2021}. Most of these methods utilize real-valued signals in the algorithms. In \cite{main_compare}, the authors use the single-sided amplitude spectrum as input to a 
CNN. However, in frequency domain, the signals are represented as complex values. Such a complex representation contains the power of a specific wave relative to others (amplitude) and the alignment of different frequency components in time (phase). Using only the amplitude information will lose the phase information, which might be critical for EEG signals. Figure~\ref{real_complex_conv_compare} illustrate the core idea of our argument. In Figure~\ref{aaaa}, we can see that these two convolution operations use the same convolution kernel, and all the complex numbers have the same modulus, which is $\sqrt{2}$; however, the convolution results are different. If we only use the modulus in this convolution operation (See Figure~\ref{bbbb}), we can only get one result, which is $4$. Suppose the phase is the only difference between two signals; using amplitude only as the feature can not differentiate them properly.

We, therefore, propose an algorithm that can utilize the features hidden in the phase information. To achieve this goal, we may train a real-valued convolutional neural network whose inputs are the amplitude and the phase as two channels or the real part and the imaginary part of the complex number as two channels. However, the traditional convolution operation contains a linear combination of the channels. It is unclear if such a linear combination is meaningful\cite{complex_patrick_2019}. 

With these issues in mind, we believe that using the original complex values of DFT as input is a better alternative. In  \cite{Hirose_generalization}, the authors also mentioned that complex-valued neural networks might be suitable for extracting phase information. Therefore, we exploit a complex-valued neural network for our case, more specifically complex-valued convolutional neural network. In \cite{Guberman_2016_on_complex}, a complex-valued convolution neural network was introduced and compared with a real-valued convolutional neural network on the classification tasks. Another notable paper is~\cite{crelu}, where the authors applied deep complex networks to audio-related tasks and achieved promising results; this work also presented batch-normalization and complex weight initialization strategies for complex-valued neural networks. In our experiments, we observe that the performance is not improved if all the layers in the network are complex-valued. The first possible reason is the complex-valued non-linear activation function is task-specific; an inadequate selection of activation function may lead to poor transmission of the information between layers\cite{complex_patrick_2019, Scardapane_complex_2018, bassey_2021_survey}. The second possible reason is that in the loss function of a complex-valued neural network, one usually needs to calculate the "distance" between complex numbers and real numbers, which is not well-defined in mathematics. 

Therefore, we develop an algorithm that integrates the real-valued and complex-valued neural networks to overcome the difficulties mentioned above. Our approach builds on the network structure of San-Segundo \textit{et~al.} \cite{main_compare}, which contains two convolutional layers for feature extraction and three fully connected layers for classification. We improve their neural network by adding a crucial complex-valued convolutional layer at the beginning, and immediately taking the modulus as non-linear activation (See Section~\ref{structure_section}). There are four main advantages of our network architecture:  1) The extra complex-valued convolutional layer captures the features in the phase information of complex-valued input; 2) we only need one universal complex-valued activation function, which is taking the modulus; 3) The difficulty in distance calculation between complex numbers and real numbers in the loss function is avoided in our framework; 4) Our network 
uses about 50\% fewer parameters compared with the structure in \cite{main_compare}. Moreover, our framework can achieve higher classification accuracies than the conventional feature selection method and the CNN in \cite{main_compare} in both the experiments using a benchmark dataset and the simulations.

We apply the proposed method to the EEG signal classification problem with discrete Fourier transform (DFT). We present qualitative results on several classification tasks, including binary and multi-class classification on two simulated datasets and one real-world dataset. The simulated datasets consist of synthetic signals of our own design and signals generated based on well-known theories. The real-world dataset is the Ralph-Andrzejak EEG dataset \cite{RalphEEG}, which contains 500 EEG recordings from 500 patients.

The rest of this paper is structured as follows: Section 2 contains the methodology in this study, including a description of the discrete Fourier transform (DFT), the framework of our neural network, the backpropagation algorithm, and some training details. Section 3 describes the experiments and the results obtained in simulation and a real-world dataset. Section 4 summaries this paper and discusses the limitations of our method and our future works

The symbols and notations used in this paper are summarized in Table~\ref{symbols_notations}
\begin{table}[!ht]
	\setlength{\tabcolsep}{10pt}
	\begin{center}
		\begin{tabular}{ c  l }
			$\mathbb{R},\ \mathbb{C}$  &   Sets of real, complex numbers \\  
			$\mi$         &   Imaginary unit            \\
			$z^\intercal,\ z^*,\ z^H$  & Transpose, conjugate, conjugate transpose of $z$ \\
			$\mathfrak{Re}(z)$ & Real part of $z$\\
			$abs(z),|z|$ & Absolute value, modulus of $z$ 
% 			$$         & Conjugate of $z$ \\ 
% 			$$         & Conjugate transpose of $z$ \\
% 			RVNN, CVNN          & Real-valued, complex-valued neural network\\
% 			BP            & Backpropagation\\
% 			$\nabla$      & Gradient operator
		\end{tabular}
	\end{center} 
	\vspace{-0.1in}
	\caption{SYMBOLS AND NOTATIONS}
	\vspace{-0.1in}
	\label{symbols_notations}
\end{table}

\section{Methodology}
In this section, we present an overview of our methodology, including the framework of our network, a description of discrete Fourier transform, the backpropagation algorithm, and some training details.

\subsection{Framework}
\label{structure_section}
\begin{figure*}[!ht]
    \centering
	\includegraphics[width=0.9\textwidth]{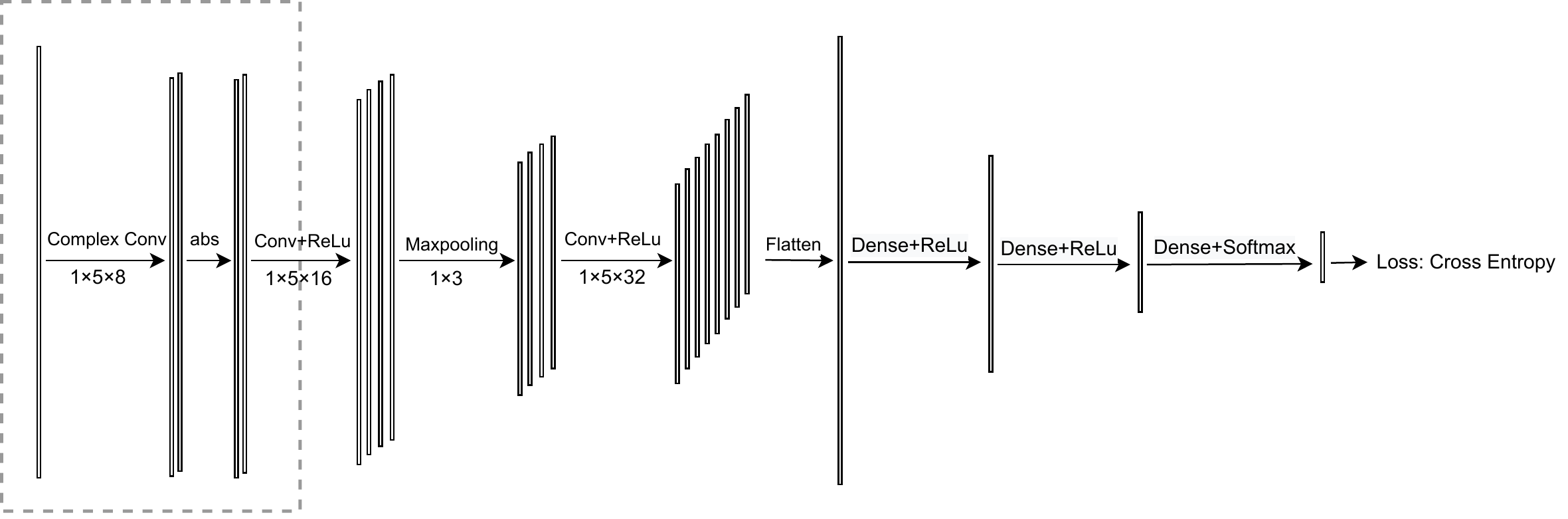}
	\vspace{-0.1in}
	\caption{Our neural network structure.}
	\label{structure}
	\vspace{-0.1in}
\end{figure*}

Figure~\ref{structure} shows our framework. The first layer we use is a complex-valued convolutional layer. Immediately after this layer, we take the modulus (see the part in the dotted line box in Figure~\ref{structure}), making all the outputs real-valued. After this, we add two real-valued convolutional layers and three fully connected layers with ReLu and max-pooling. Finally, we use softmax as the last activation function and cross-entropy as the loss function. 

\subsection{Discrete Fourier Transform and Its Inverse Transform}
We apply discrete Fourier transform (DFT) on the original EEG signals to obtain their representations on the frequency domain and apply inverse discrete Fourier transform (IDFT) to achieve the inverse transform. The formulas of DFT and IDFT are shown below:
\begin{align}
    DFT:\ \ \ &\hat{x}(k) = \sum_{j=1}^nx(j)\omega_n^{(j-1)(k-1)}, \ k = 1\ ...\ n  \label{fft_formula}\\
    IDFT:\ \ \ &x(j) = \frac{1}{n}\sum_{k=1}^n\hat{x}(k)\omega_n^{-(j-1)(k-1)}, \ j = 1\ ...\ n  \label{ifft_formula}
\end{align}
where $x$ is the original signal of length $n$, $\hat{x}$ is the Fourier transform of $x$ of length $n$, and $\omega_n = e^{-2\pi \mi/n}$. This paper implements the DFT and IDFT using MATLAB command $fft()$ and $ifft()$ . Because our original signals are all real-valued, their DFTs are conjugate symmetric. We only keep the first half part of the DFTs as the input to our neural network.
\subsection{Backpropagation}
As we can see from Figure~\ref{structure}, our framework contains a complex-valued convolutional layer taking modulus as activation, and after this, all the layers are real-valued. We use the Adam algorithm \cite{adam} with default settings to optimize the kernel and bias in real-valued layers. To optimize the parameters in the complex-valued convolutional layer, we also need the backpropagation algorithm in the complex value. The regular complex derivative only applies to the analytic functions \cite{Scardapane_complex_2018}; however, to obtain the modulus need to evaluate the following the function:
\begin{equation}
    f(Z) = |Z| =  (Z^*Z)^{\frac{1}{2}}
    \label{absolute}
\end{equation}
which is not analytic. So in backpropagation, the derivative of $f(Z)$ with respect to $Z$ can not be calculated with a regular complex derivative. In this case, we need to adopt the Wirtinger derivative.

\subsubsection{Wirtinger derivative}
The following defines the Wirtinger derivatives: 
\begin{definition}
Consider the complex plane $\mathbb{C} \equiv \mathbb{R}^2=\{(x,y)\ |\ x,y\in\mathbb{R}\}$. The two operators $\frac{\partial}{\partial z}$ and $\frac{\partial}{\partial z^*}$ are defined by:

\begin{equation}
    \begin{split}
        \frac{\partial}{\partial z}      := & \frac{1}{2}\left[\frac{\partial}{\partial x}-\mi \frac{\partial}{\partial y}\right],\\
        \frac{\partial}{\partial z^*}    := & \frac{1}{2}\left[\frac{\partial}{\partial x}+\mi \frac{\partial}{\partial y}\right].
    \end{split}
    \label{wirtinger}
\end{equation}

are referred to as the Wirtinger derivatives~\cite{wirtinger_zur_1927}.
\label{def_wirtinger}
\end{definition}

Wirtinger derivative holds standard rules for differentiation known from real-valued analysis concerning the sum, product, and composition of two functions. Then from equation~(\ref{wirtinger}), we can derive another essential property of Wirtinger derivative:
\[ \frac{\partial z}{\partial z^*}=0 \ , \ \ \ \ \frac{\partial z^*}{\partial z}=0, \]
which means we can treat $z$ and $z^*$ as independent variables. Then based on the first-order Taylor expansion for multivariable functions, we have:
\begin{equation}
    df = \frac{\partial f}{\partial z}dz + \frac{\partial f}{\partial z^*}dz^* + \mathcal{O}(|dz|^2)
    \label{first_order_taylor_expansion}
\end{equation}

To further derive the backpropagation algorithm based on the Wirtinger derivative, we need the following Corollary.
\begin{corollary}
Derivatives of the conjugate function $f^*(z)$ satisfy the following relationships:
\begin{equation}
    \frac{\partial f^*(z)}{\partial z}   = \left( \frac{\partial f(z)}{\partial  z^* } \right)^* \ \ \ \ \ \ \ \ \ 
    \frac{\partial f^*(z)}{\partial z^*} = \left( \frac{\partial f(z)}{\partial  z } \right)^*.
    \label{coro_1_equ}
\end{equation}
\label{corollary_1}
\end{corollary}
It is straightforward to derive equation~(\ref{coro_1_equ}) from Definition~\ref{def_wirtinger}.

In our case,  $f(z):\mathbb{C}\rightarrow \mathbb{R}$, which means $f(z)=f^*(z)$, equation~(\ref{coro_1_equ}) can be further simplified to the following equations:
\begin{equation}
    \frac{\partial f(z)}{\partial z}   = \left( \frac{\partial f(z)}{\partial  z^* } \right)^* \ \ \ \ \ \ \ \ \ 
    \frac{\partial f(z)}{\partial z^*} = \left( \frac{\partial f(z)}{\partial  z } \right)^*.
    \label{coro_real_case_equ}
\end{equation}
From equations~(\ref{first_order_taylor_expansion}) and~(\ref{coro_real_case_equ}), by omitting the lower order term, we then have:
\begin{equation}
    \begin{split}
        df &= \frac{\partial f}{\partial z}dz + \frac{\partial f^*}{\partial z^*}dz^*\\
          &= \frac{\partial f}{\partial z}dz + \left(\frac{\partial f}{\partial z}dz\right)^*\\
          &= 2\mathfrak{Re}\left(\frac{\partial f}{\partial z}dz\right)
    \end{split}
\end{equation}
Based on the principles of the gradient descent method, we need to find $dz$ that maximize $\left|2\mathfrak{Re}\left(\frac{\partial f}{\partial z}dz\right)\right|$. From Cauchy–Schwarz inequality, we know that $dz$ should have the same direction of $\left(\frac{\partial f}{\partial z}\right)^*$. So the direction of the steepest ascent is the direction of $(\frac{\partial f}{\partial z})^*$(if $f(z):\mathbb{C}\rightarrow \mathbb{R}$, from Equation~\ref{coro_real_case_equ}, we know that $\frac{\partial f}{\partial z^*}=(\frac{\partial f}{\partial  z })^*$).

We can give the general form of the backpropagation algorithm for CVNN:
\begin{equation}
    W^{(t+1)} = W^{(t)} - \eta\left(\frac{\partial Loss(W^{(t)})}{\partial W^*}\right)^\intercal
\end{equation}
where $W^{(t)}$ is the set of all the parameters that need to be learned in the complex-valued layers at $t$'th iteration. $\eta$ is the learning rate. We adopt automatic differentiation in Tensorflow \cite{tensorflow2015-whitepaper} to calculate the gradients for complex-valued layers based on the Wirtinger derivative \cite{tensorflow2015-whitepaper}. We use complex Adam \cite{complex_sarroff_2018} as the specific backpropagation algorithm to optimize the parameters. 

\subsubsection{An example of one convolution kernel}
\begin{figure}
    \centering
	\includegraphics[scale=0.8]{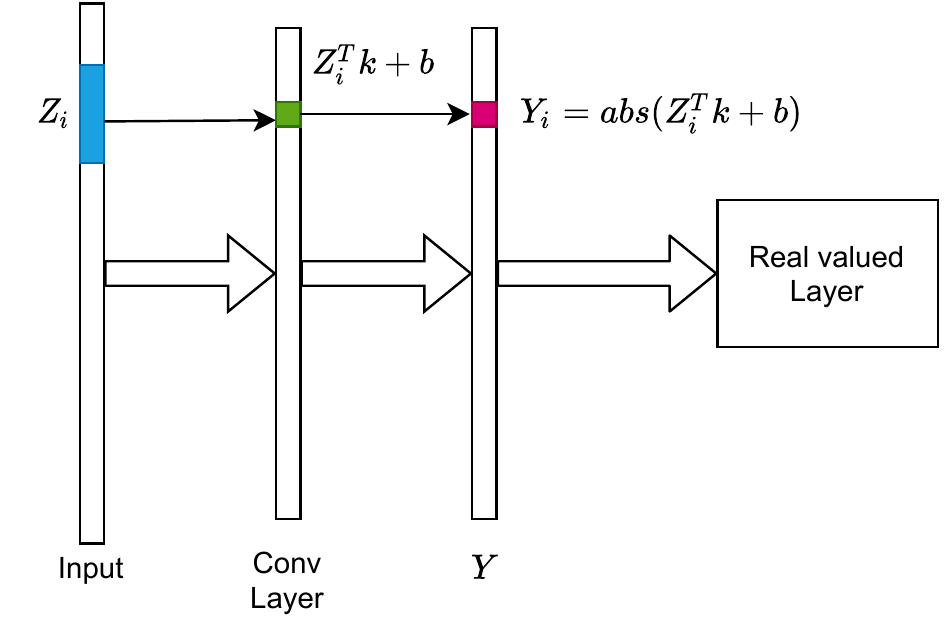}
	\vspace{-0.1in}
	\caption{An example of the forward propagation in the complex-valued convolutional layer. $Z_i$: a part of the input, $k$: complex-valued convolution kernel, $b$: complex-valued bias, $Y_i$: the modulus of the output of the complex-valued convolutional layer, $Y$: the vector whose $i$'th entry is $Y_i$. $Y$ is the input to the next real-valued layer.}
	\label{structure_1}
	\vspace{-0.1in}
\end{figure}

Suppose we have an input sequence data $Z$ and a complex convolution kernel $k$ (See Figure~\ref{structure_1}). $Z_i$ is an arbitrary part of $Z$, and $Z_i^\intercal k$ is the convolution result. Note that here $Z_i^\intercal k$ is not the Hermitian inner product of two complex-valued vectors. After adding the bias term $b$, $Y_i$ can be defined as:
\begin{equation}
    Y_i = abs(Z_i^\intercal k+b) = [(Z_i^\intercal k+b)^*(Z_i^\intercal k+b)]^{\frac{1}{2}}
    \label{aaa}
\end{equation}
From the backpropagation for the real-valued CNN, we can get $\partial Loss/\partial Y_i$. Then based on the Wirtinger derivative, we need to find $\partial Loss/\partial k^*$ and $\partial Loss/\partial b^*$. Based on the chain rule, we have:
\begin{equation}
    \frac{\partial Loss}{\partial k^*} = \sum_i \frac{\partial Loss}{\partial Y_i} \frac{\partial Y_i}{\partial k^*} = \frac{1}{2} \sum_i \frac{\partial Loss}{\partial Y_i}  (Z_i^\intercal k+b)Z_i^H
    \label{dLdk}
\end{equation}
Similarly,
\begin{equation}
    \frac{\partial Loss}{\partial b^*} = \sum_i \frac{\partial Loss}{\partial Y_i} \frac{\partial Y_i}{\partial b^*} = \frac{1}{2} \sum_i \frac{\partial Loss}{\partial Y_i}  (Z_i^\intercal k+b)
    \label{dLdb}
\end{equation}

Based on (\ref{dLdk}) and (\ref{dLdb}), we can finally find the proper gradients for $k$ and $b$.

\subsection{Other training details}
For the real-valued weights and bias, we use Xaiver initialization  \cite{xavier}. For the complex-valued weights and bias, we use the Rayleigh distribution to generate the modulus of the complex number ($r$) and the Uniform distribution $(U_{[-\pi,\pi]})$ to generate the angle ($\theta$). Then we can get the initialization for complex-valued parameters by using the formula $re^{\mi\theta}$. 

\section{Experiments}
In this section, we compare our method against two existing frameworks on the classification task with two simulated datasets. We then apply our approach to a real-world dataset and compare it with several previous works.
\subsection{Simulation Study}
In this study, we simulate EEG signals with two different methods to compare the classification performance between our method and other methods. In the first simulation, we adopt the first-order autoregression (AR(1)) model to generate the amplitude and phase separately. We then use the inverse discrete Fourier transform (IDFT) to obtain the signals on the time domain. The main difference among signals in different classes in this simulation is the phase. We want to use this simulation to prove that our algorithm can efficiently utilize the features in phase. In the second simulation, we adopt classical theory and phase resetting theory to generate event-related potential (ERP) with noise \cite{yeung_detection_2004,yeung_theta_2007}, and then we design four classification task (See Table~\ref{bc_table}). There are two reasons we perform the second simulation: 1) the signals in the second simulation are closer to the real EEG signals, 2) ERP-signal classification is crucial in analyzing human EEG activities and can be a promising tool to study error detection and observational-learning mechanisms in performance monitoring \cite{vasios_classification_2009}.

We mainly compare our method with real-valued CNN and the conventional feature selection method in these two classification tasks. For real-valued CNN, the network structure we choose to compare with is the structure used in \cite{main_compare}, which contains two convolutional layers and three fully connected layers. To compare with the feature selection method, we choose seven features, which are Shannon entropy \cite{ibrahim_electroencephalography_2018}, Renyi entropy \cite{das_discrimination_2016}, log-energy entropy \cite{das_discrimination_2016}, approximate entropy \cite{approximate_sohn_2006}, sample entropy \cite{richman_physiological_2000, li_assessing_2015}, fuzzy entropy \cite{simons_fuzzy_2018, chen_measuring_2009, shi_entropy_2017}, and exponential energy \cite{o.k._time-domain_2019}(See Table~\ref{parameters} for detailed parameter settings.). In both simulations, we applied 6th order Butterworth low pass filter to remove the frequencies over 60 Hz before extracting the features. We try all 127 combinations of these seven features. For each combination of the features, we extract them from the pre-processed signal, the first and second order difference of the pre-processed signal. So the number of features we select is always a multiple of 3. The classifier we choose is the support vector machine(SVM) for the binary classification task and the error-correcting output codes (ECOC) model using SVM binary learners\cite{escalera_decoding_2010,escalera_separability_2009} for the multi-class classification task. 

\begin{table}[!ht]
\scriptsize
	\setlength{\tabcolsep}{4pt}
	\begin{center}
		\begin{tabular}{ c | c | c}
			Features    & Parameters & Ref.           \\ 
			\hline
			Renyi Entropy & $\alpha=0.5$ & --\\ 
			Approximate Entropy & $m=2,\tau=1,r=0.2*sd$&\cite{richman_physiological_2000} \\
			Sample Entropy& $m=2,\tau=1,r=0.2*sd$ &\cite{richman_physiological_2000} \\   
			Fuzzy Entropy & $m=3,\tau=3,r=0.15*sd$&\cite{li_detection_2018} \\ 
		\end{tabular}
	\end{center} 
	\vspace{-0.1in}
	\caption{Parameter details. $\alpha$: order of Renyi entropy, $m$: embedding dimension, $\tau$: time delay, $r$: threshold value to determine similarity, $sd$: the standard deviation of the input time-series data.}
	\label{parameters}
	\vspace{-0.1in}
\end{table}

In this simulation study, we use 5-fold cross-validation to obtain accuracy. To alleviate the accuracy variation, we repeat the 5-fold cross-validation ten times for each classification task to get the average and standard deviation of the accuracies. The accuracies presented in this simulation study are based on the results with the highest mean accuracy.

\subsubsection{EEG signals generated with AR(1) model}

In this section, we simulate the signals using AR(1) model because we assume that neighboring amplitude and phase are not independent. The AR(1) model assumes that the current value depends linearly on its immediately prior value and a stochastic term, which complies with our assumptions. The formula for AR(1) model is shown in Equation~(\ref{ar1}):
\vspace{-0.01in}
\begin{equation}
    x_t = \beta_0 + \beta_1 x_{t-1} + \epsilon_t
    \label{ar1}
\vspace{-0.01in}    
\end{equation}
where $x_t$ is the present value, $x_{t-1}$ is the immediately prior value, $\beta_0$ is a constant, $\beta_1$ is the model parameter, and $\epsilon_t$ is the white noise with zero mean and constant variance.

\begin{figure}[!t]
    \centering
	\includegraphics[width=0.9\textwidth]{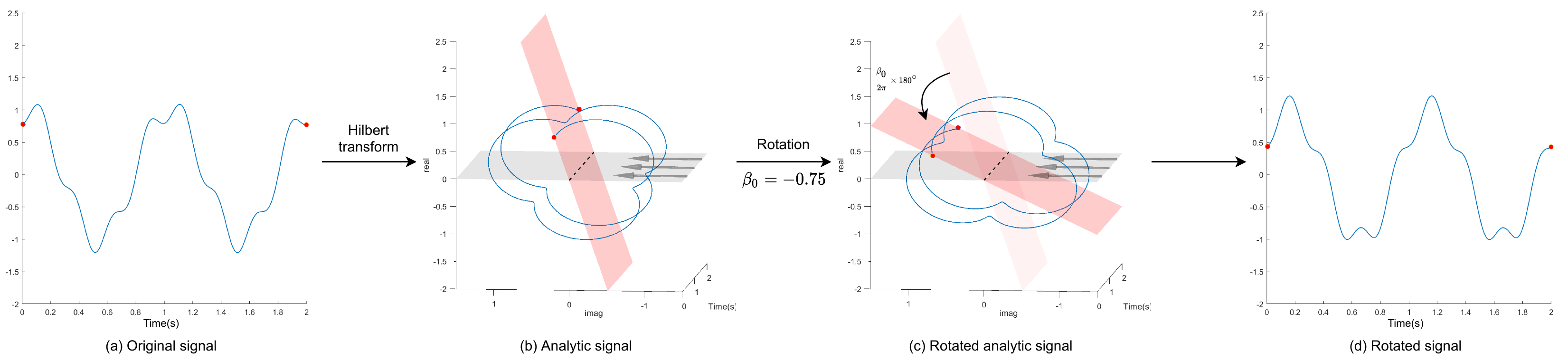}
	\vspace{-0.1in}
	\caption{The effect of $\beta_0$: (a) the original signal, (b) the analytic signal obatined by applying Hilbert transform on the original signal. (If we observe the analytic signal (b) in the direction of the arrow, we can see the original signal in (a).) (c) the rotation of the analytic signal by $\beta_0$. (d) the rotated signal (If we observe the rotated analytic signal (c) in the direction of the arrow, we can see the signal in (d).)}
	\label{hilbert_rot}
	\vspace{0.2in}
\end{figure}

We first simulate the phase information. We know that the phase $\theta \in [-\pi, \pi]$, so we modify the Equation~(\ref{ar1}) to ensure the range of $\theta$ is limited. The modified formula is shown in Equation~(\ref{mod_ar1}):
\vspace{-0.01in}
\begin{equation}
    \theta_t = Rem(\beta_0 + \beta_1\theta_{t-1} + \epsilon_t, \pi)
    \label{mod_ar1}
    \vspace{-0.01in}
\end{equation}
where $Rem(\star,\pi)$ is a function used to obtain the remainder of $\star$ divided by $\pi$. Here, $\beta_0$ achieves an overall phase shift, and its effect can be understood as a rotation of a signal under the Hilbert transform (See Figure~\ref{hilbert_rot}). Suppose we have the Hilbert transform of a real-valued signal. In that case, we can plot the analytic form of the signal in three-dimensional Cartesian coordinates with the time axis, the real part axis, and the imaginary part axis. Then we can rotate this analytic form of the signal about the time axis(dashed line in Figure~\ref{hilbert_rot} (b), (c)) to achieve the effect of $\beta_0$. 

\begin{figure}[!t]
    \centering
	\includegraphics[width=0.9\textwidth]{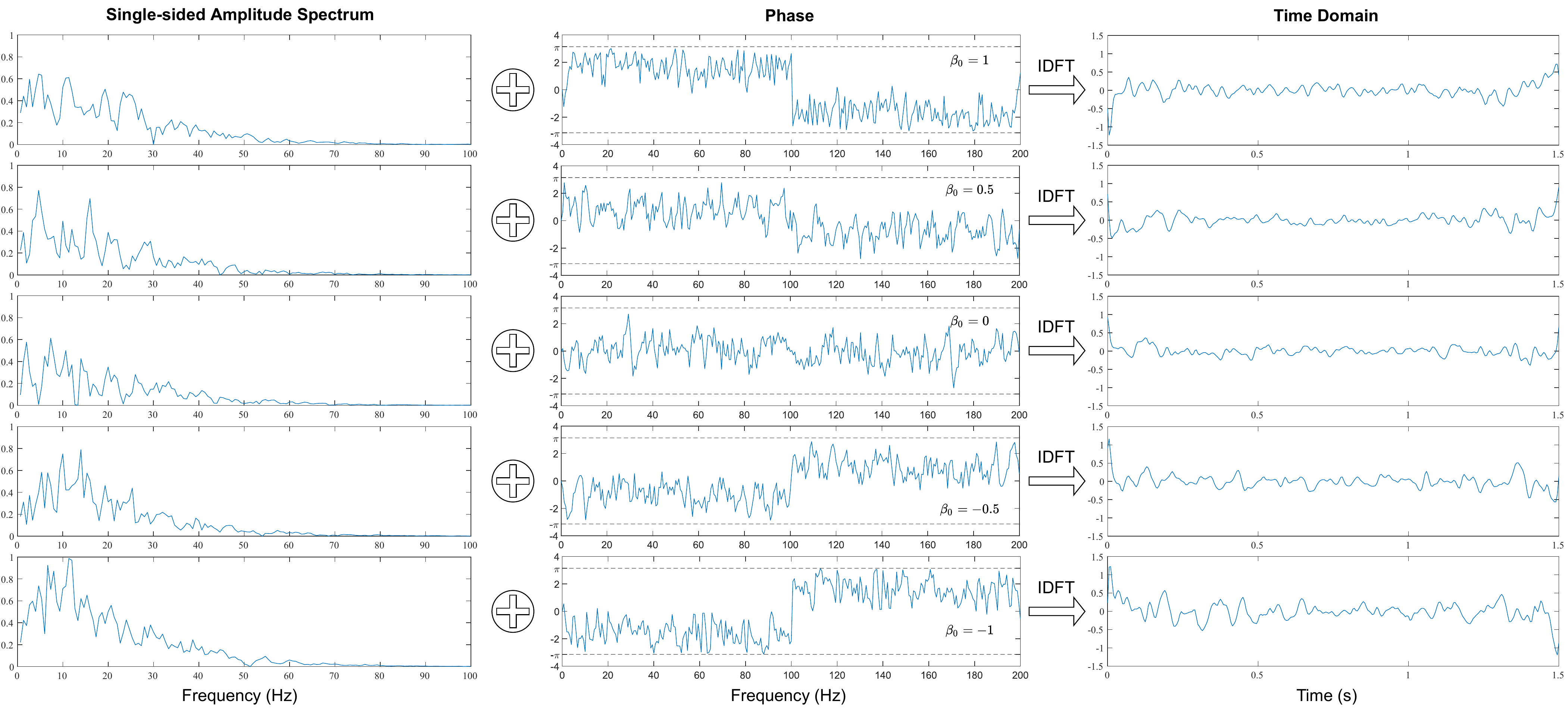}
	\vspace{-0.1in}
	\caption{The first column shows the single-sided amplitude spectrum generated using AR(1) model with $\beta_0=0, \beta_1=0.5, Var(\epsilon_t)=0.5$, the second column shows the simulated phase information generated using modified AR(1) formula with $\beta_0=0,\pm0.5,\pm1, \beta_1=0.5, Var(\epsilon_t)=0.5$ and the last column shows the simulated signals generated from the IDFT of the amplitude spectrum and phase.}
	\label{sim_signal}
	\vspace{0.05in}
\end{figure}

\begin{figure}[!t]
     \centering
     %%%%%%%%%%%%%%%%%%%%%%%%%%%%%%%%%%%%%%%%%
     \begin{subfigure}[t]{0.44\textwidth}
         \centering
         \caption{CNN (phase)}
         \vspace{-0.05in}
         \includegraphics[width=\textwidth]{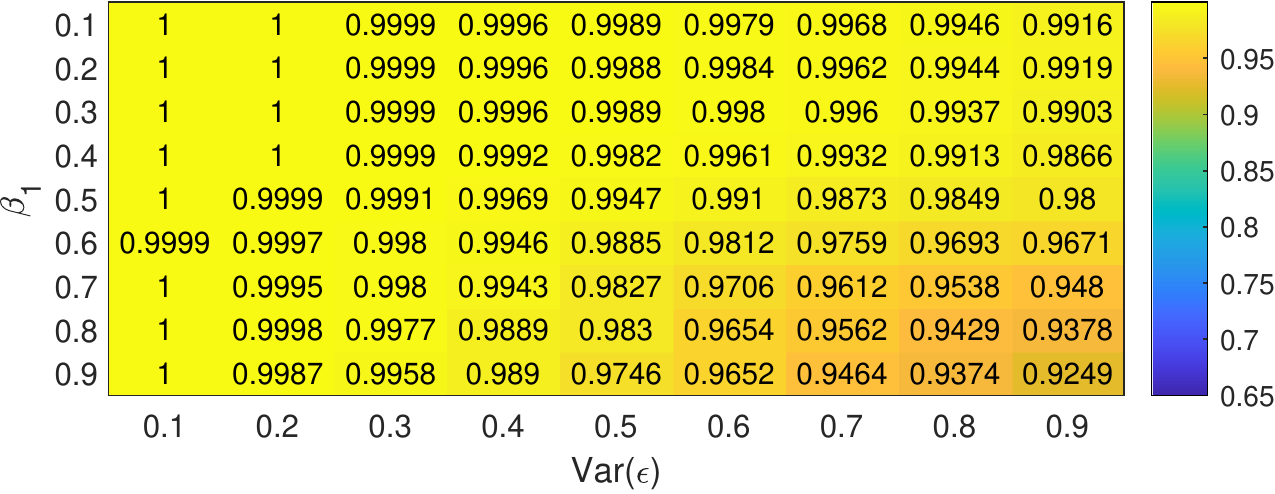}
         \label{cnn_phase}
     \end{subfigure}
    %  \hfill
     \begin{subfigure}[t]{0.44\textwidth}
         \centering
         \caption{CNN (time domain)}
         \vspace{-0.05in}
         \includegraphics[width=\textwidth]{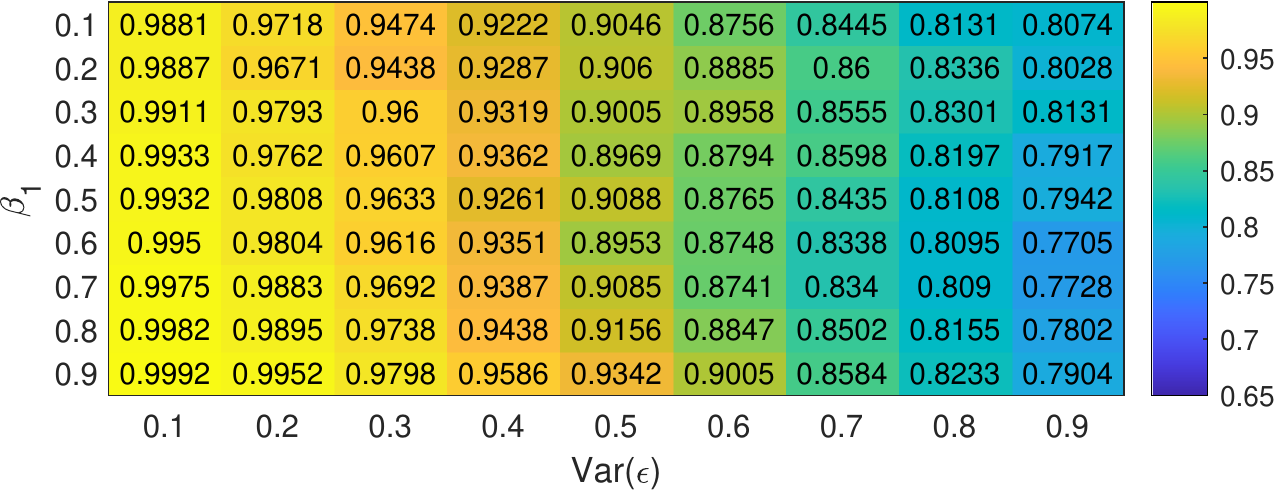}
         \label{cnn_time}
     \end{subfigure}
     
    %  \vspace{-0.1in}
    \begin{subfigure}[t]{0.44\textwidth}
         \centering
         \caption{Feature selection method (time domain)}
         \vspace{-0.05in}
         \includegraphics[width=\textwidth]{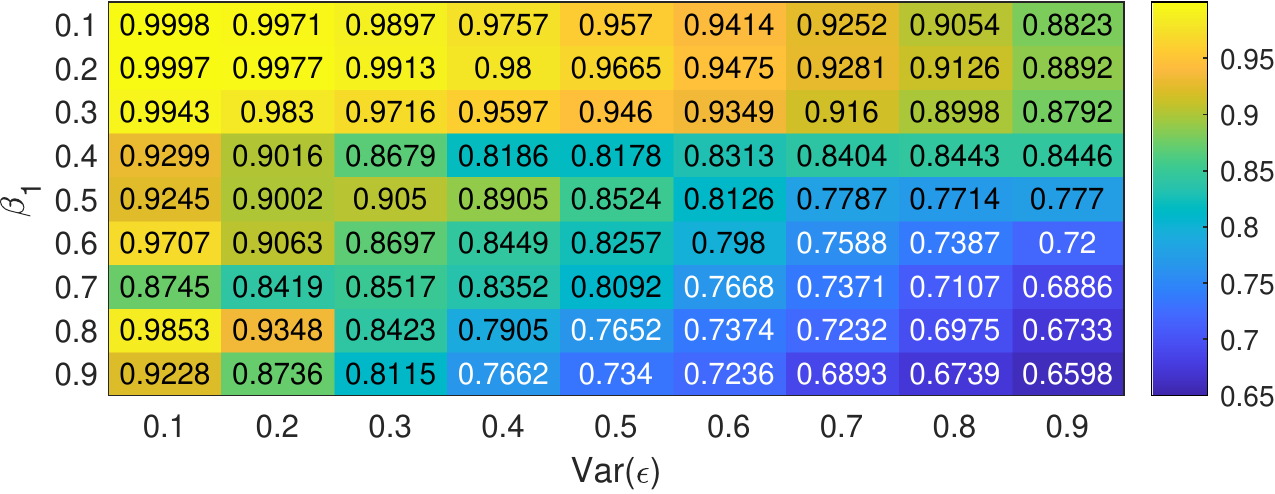}
         \label{traditional_time}
     \end{subfigure}
    %  \hfill
     \begin{subfigure}[t]{0.44\textwidth}
         \centering
         \caption{Our method (frequency domain)}
         \vspace{-0.05in}
         \includegraphics[width=\textwidth]{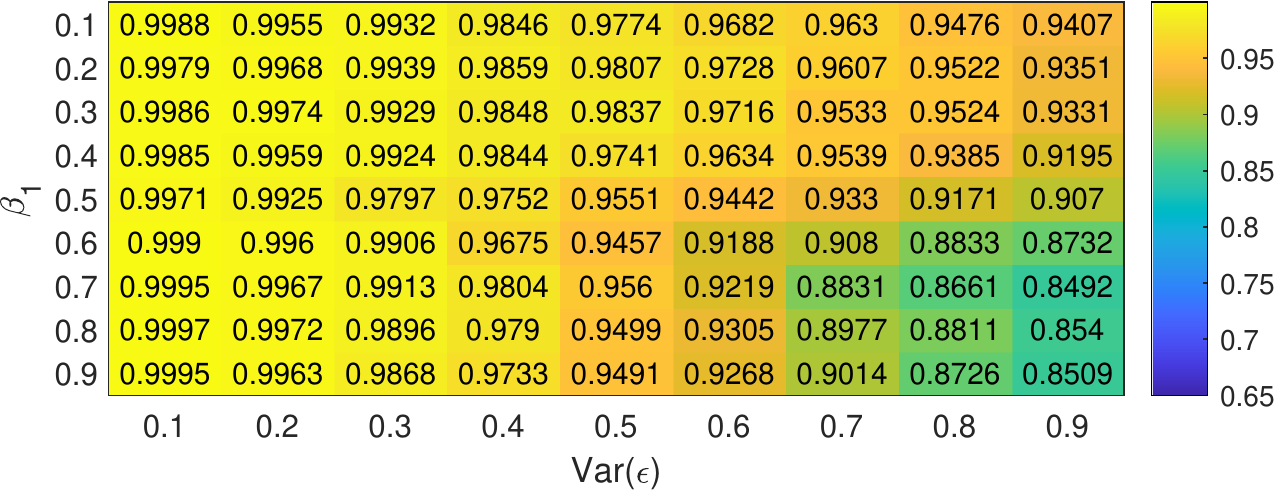}
         \label{our_fre}
     \end{subfigure}
     \vspace{-0.2in}
     \caption{(a) the accuracies of using CNN with phase information only, (b) the accuracies of using the CNN with the input of simulated signals (time domain), (c) the accuracies of applying the feature selection method, (d) the accuracies of using our method on simulated signals (frequency domain).}
     \label{acc2}
\end{figure}

From equation~(\ref{mod_ar1}), we can see that, to simulate the phase, we need to determine three parameters \textemdash $\ \beta_0,\ \beta_1$, and the variance of $\epsilon_t$. Since $\beta_1$ determines the correlation between the $\theta_t$ and $\theta_{t-1}$ and the variance of $\epsilon_t$ determines the intensity of the noise, we use different $\beta_0$ as the baseline to generate signals for different classes (See the middle column in Figure~\ref{sim_signal}). Because we want to compare our method with other methods in multi-class classification, we generate signals for five classes by setting $\beta_0$ as $0,\pm0.5,\pm1$ (corresponds to the different angle of rotation in Figure~\ref{hilbert_rot}). To show the effect of $\beta_1$ and the variance of $\epsilon_t$ for different classes, we let $\beta_1$ and $Var(\epsilon_t)$ be ranged from 0.1 to 0.9 with an interval of 0.1. The accuracy table in Figure~\ref{acc2} contains nine by nine values, and each value corresponds to a specific pair of $\beta_1$ and $Var(\epsilon_t)$. 

\begin{figure}[!t]
    \centering
	\includegraphics[width=0.9\textwidth]{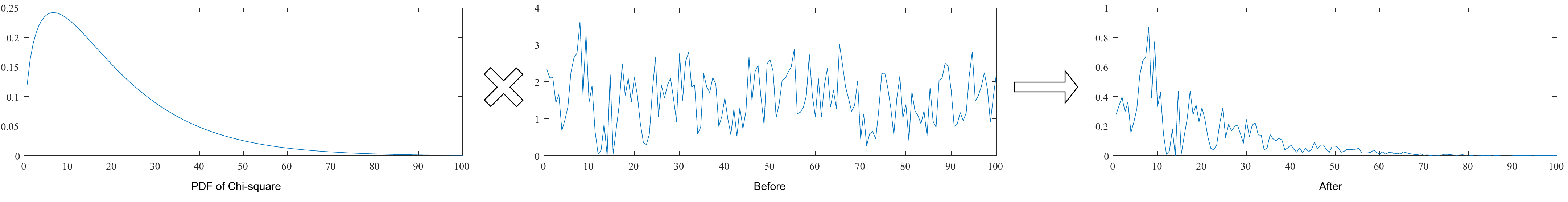}
	\vspace{-0.1in}
	\caption{Simulated single-sided amplitude spectrum: the left graph shows the probability density function of the Chi-squared distribution, the middle graph is the single-sided amplitude spectrum generated with the unmodified AR(1) model, and the right graph shows the single-sided amplitude spectrum multiplied with the probability density function of the chi-squared distribution.}
	\label{sim_chi}
%  	\vspace{0.2in}
\end{figure}

After we obtain the simulated phase, we use unmodified AR(1) model (\ref{ar1}) with $\beta_0=0, \beta_1=0.5, Var(\epsilon_t)=0.5$ to randomly generate the amplitude for different groups. We then multiplied the single-sided amplitude spectrum by the Chi-square distribution to make our simulated signals have the main bandwidth appears in the range of 0 to 70 Hz (see Figure~\ref{sim_chi}), which is close to the bandwidth used by clinical analysis of EEG \cite{beniczky_electroencephalography_2020}. Then we adopt IDFT to obtain the simulated signals on the time domain using the simulated phase and amplitude (see Figure~\ref{sim_signal}). Finally, the simulated signals last for 1.5 seconds with a 200 Hz sampling frequency.


\begin{figure}[!th]
    \centering
    \includegraphics[width=0.9\textwidth]{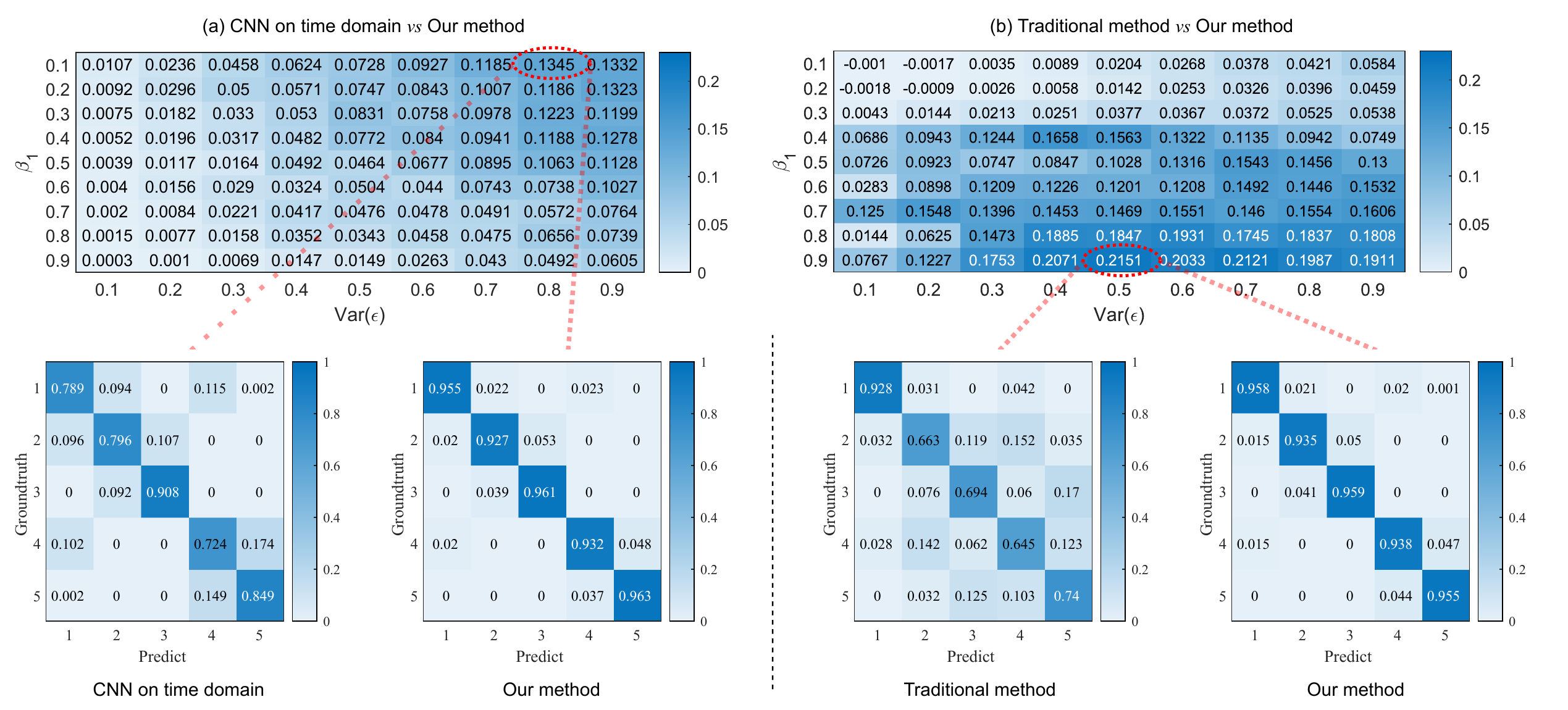}
    \vspace{-0.1in}
    \caption{(a): the accuracy differences between our method and applying CNN on the time domain. (b): the accuracy differences between our method and the traditional method. From (a), we can see that when $Var(\epsilon)=0.8$ and $\beta_1=0.1$, our method can outperform CNN on the time domain the most. As shown in (b), our method can outperform traditional method the most when $Var(\epsilon)=0.5$ and $\beta_1=0.9$. The bottom figures are the confusion matrices when the accuracy difference achieves the largest (labels 1 to 5 correspond to $\beta_0=0,\ 0.5,\ 1,\ -0.5,\ -1$, respectively).}
    \label{diff_confmat}
\end{figure}

 Figure~\ref{acc2} shows the classification results using this simulated dataset. As shown in Figure~\ref{cnn_phase}, it is not surprising that applying CNN on phase-only data can achieve the highest accuracy no matter the value of $\beta_1$ and $Var(\epsilon)$. Because based on our simulation, all the differences among different classes are reflected in the phase information, and the random variation in amplitude can be viewed as noise. We also apply CNN and the feature selection method to the simulated signals in the time domain for comparison (see Figure~\ref{cnn_time} and~\ref{traditional_time}). Figure~\ref{our_fre} shows the result obtained by using our algorithm. Figure~\ref{diff_confmat} shows the accuracy differences between our method and the two methods for comparison. We can see our method outperforms other methods in most cases.  We also show the confusion matrices in the lower part of Figure~\ref{diff_confmat}. These four confusion matrices are selected when the accuracy difference achieves the largest($Var(\epsilon)=0.8,\ \beta_1=0.1$ and $Var(\epsilon)=0.5, \ \beta_1=0.9$). 

\subsubsection{EEG signals generated according to classical theory and phase resetting theory}
In this simulation, we generate EEG signals for the classification task according to two main theories: classical theory and phase-resetting theory \cite{yeung_detection_2004,yeung_theta_2007}. The former theory assumes that the ERP signal is buried in the ongoing EEG noise, while the latter theory believes that the events can reset the phase of ongoing oscillations. With each theory, we design two binary classification tasks, referred to as the "Fixed location" task and the "Random location" task (see Table~\ref{bc_table}), to compare our method and other methods.

\begin{table}[!h]
\scriptsize
	\begin{center}
		\begin{tabular}{c | c | c}
		    & classical theory & phase resetting theory \\
		    \hline
		    \hline
		    Fixed location&Fixed peak location + noise $\bm{vs}$ noise only& Fixed phase resetting location + noise $\bm{vs}$ no phase resetting  + noise\\
		    \hline
		    Random location&Random peak location + noise $\bm{vs}$ noise only&random phase resetting location  + noise $\bm{vs}$ no phase resetting  + noise
		\end{tabular}
	\end{center} 
	\vspace{-0.1in}
	\caption{The four binary classification tasks we used to compare our approach and other methods.}
	\label{bc_table}
% 	\vspace{-0.1in}
\end{table}

\begin{figure}[!h]
    \centering
	\includegraphics[width=0.9\textwidth]{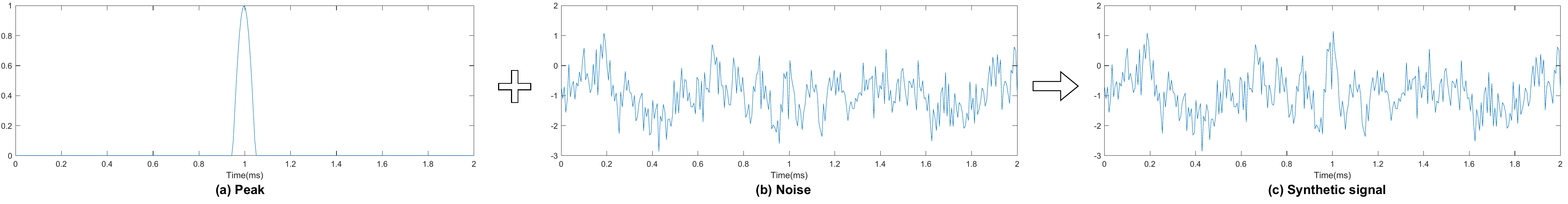}
	\vspace{-0.1in}
	\caption{An example of the peak, noise, and synthetic signal generated according to classical theory. (a) Peak: the highest value of the peak signal shows exactly at 1s with the frequency of 5 Hz. (b) Noise: Randomly generated based on human EEG background signal spectrum. (c) Synthetic signal: the weighted summation of peak signal and noise signal.}
	\label{noise_peak}
\end{figure}

In the simulation based on the classical theory, we first randomly generated 10,000 pieces of noise signals with the same power spectrum of human EEG signals using the MATLAB codes downloaded from \cite{EEG_sim_website}. Each piece of noise signal lasts for 2 seconds with a sampling rate of 150 Hz. Then 5000 pieces of these noise signals were randomly selected to add a peak signal. In the experiment of the "Fixed location," we added 5 Hz peak signal centered at the middle of each piece of noise (see Figure~\ref{noise_peak}), and in the experiment of the "Random location," we change the location of the center of the peak to be uniformly random distribute on the interval of 0 to 2 seconds.
 
\begin{figure}[!t]
    \centering
	\includegraphics[width=0.9\textwidth]{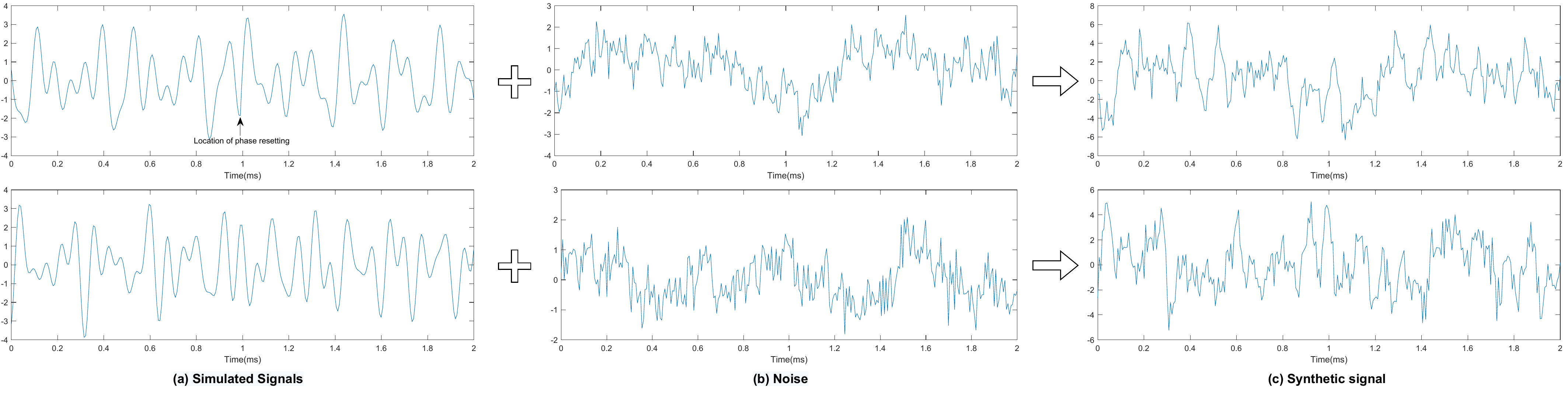}
	\vspace{-0.1in}
	\caption{An example of the signals generated according to phase resetting theory. (a) an example of simulated signals with (top) and without (bottom) phase resetting. (b) Randomly generated noise based on human EEG background signal spectrum. (c) the synthetic signals, which are the weighted summation of signals in (a) and (b).}
	\label{pr_noise}
\end{figure}

In the simulation guided by phase resetting theory, we randomly generate 5000 signals with phase resetting and 5000 signals without phase resetting. For the phase resetting group, similar to Makinen \textit{et~al.} \cite{makinen_auditory_2005}, we generate simulated data by summing four sinusoids with frequencies chosen randomly from the range 4 to 16 Hz. In the experiment of "Fixed position," each of these four sinusoids contains phase resetting at the center (see Figure~\ref{pr_noise}). In the experiment of "Random position," each of these four sinusoids contains phase resetting at the same random position (uniformly distributed on the interval of 0 to 2 seconds). Then we add randomly generated human EEG background noise to all the signals.

We mainly compare our method with the approach that applies CNN on phase information only, amplitude information only, and the original signal. We also separated the real and imaginary parts of the FFT and applied the exact structure of CNN to each of them. Furthermore, we also apply CNN with the input of real and imaginary parts as two channels. The result shows in Table~\ref{sim_2_acc}. We can see that our method can outperform others by comparing the accuracy.

\begin{table*}
\scriptsize
	\begin{center}
		\begin{tabular}{c| c || l | c | c}
		    &&Method&classical theory & Phase resetting(with noise) \\
		    \hline
		    \hline
		    \multirow{8}{*}{Fixed}
		    &Traditional&Features + classifier&0.6464$\pm$0.0010&0.6403$\pm$0.0012\\ 
		    \cline{2-5}
			     &&Phase only         &0.9758$\pm$0.0018&0.9410$\pm$0.0042\\
			     &&Amplitude only     &0.9422$\pm$0.0023&0.8640$\pm$0.0113\\
			     &&Original signal    &0.9222$\pm$0.0022&0.9025$\pm$0.0021\\
		    	&CNN&Real part only     &0.9801$\pm$0.0013&0.9113$\pm$0.0028\\
			     &&Imaginary part only&0.6349$\pm$0.0096&0.8553$\pm$0.0033\\
			     &&Real+Imaginary(two channels)     &0.9593$\pm$0.0049&0.9305$\pm$0.0017\\
			     &&Our method         &\textbf{0.9953$\pm$0.0009}&\textbf{0.9679$\pm$0.0103}\\
			\hline
			\hline
			\multirow{8}{*}{Random}
			&Traditional&Features + classifier&0.6504$\pm$0.0010&0.6807$\pm$0.0010\\
			\cline{2-5}
			      &&Phase only         &0.9273$\pm$0.0029&0.9043$\pm$0.0022\\
			      &&Amplitude only     &0.8923$\pm$0.0056&0.7629$\pm$0.0086\\
			      &&Original signal    &0.8727$\pm$0.0071&0.5371$\pm$0.0049\\
		    	&CNN&Real part only     &0.9488$\pm$0.0020&0.8840$\pm$0.0070\\
			     &&Imaginary part only&0.9431$\pm$0.0027&0.8690$\pm$0.0067\\
			     &&Real+Imaginary(two channels)     &0.9162$\pm$0.0113&0.7926$\pm$0.0067\\
			     &&Our method         &\textbf{0.9930$\pm$0.0012}&\textbf{0.9369$\pm$0.0144}\\
		\end{tabular}
	\end{center} 
	\vspace{-0.1in}
	\caption{Accuracy comparison. }
	\label{sim_2_acc}
% 	\vspace{-0.1in}
\end{table*}
\pagebreak

\subsection{Real-World Data}
\subsubsection{Data Acquisition, Description and Pre-processing}
The real-world dataset analyzed here is the Ralph-Andrzejak EEG dataset \cite{RalphEEG}, which contains five categories of signals named Z, O, F, N and S, respectively (see Table~\ref{5cluster}). Each category contains 100 EEG records of about 23.6 seconds with a sampling rate of 173.61 Hz (4097 values per record). Although there are 5 classes in this dataset, the four most common classification tasks for this dataset are Z vs S \cite{o.k._time-domain_2019}, S vs NS (NS=Z+O+F+N) \cite{wang_automatic_2017,kumar_classification_2015}, Z vs N vs S \cite{Abualsaud_ensemble_2015} and Z vs F vs S \cite{Sadati_2006}. 

\begin{table}[!ht]
\scriptsize
	\begin{center}
		\begin{tabular}{ c | l }
			Category    & Description                    \\ 
			\hline \hline
			Z           & Healthy group, recorded with eye open  \\  
			O           & Healthy group, recorded with eye closed  \\   
			F           & Interictal activity, recorded in the epileptogenic zone \\ 
			N           & Interictal activity, recorded at hippocampal location      \\  
			S           & Seizure activity  \\   
		\end{tabular}
	\end{center} 
	\vspace{-0.2in}
	\caption{Detailed description of the 5 categories of EEG signals in the Ralph-Andrzejak EEG dataset}
	\label{5cluster}
% 	\vspace{-0.1in}
\end{table}

\begin{figure*}[!t]
    %\centering
    \begin{center}
	\includegraphics[width=0.9\textwidth]{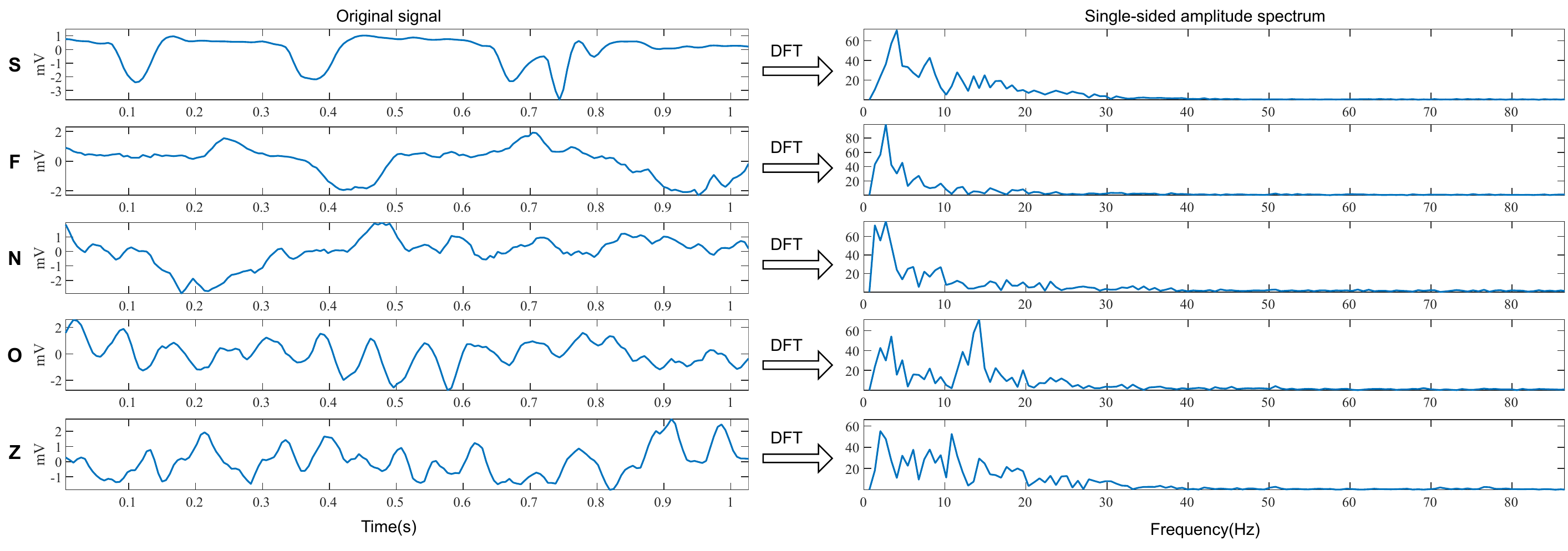}
	\vspace{-0.1in}
	\end{center}
	\vspace{-0.2in}
	\caption{Five example signals in Ralph-Andrzejak EEG dataset (left) and their single-sided amplitude spectrum (right).}
	\label{fft}
% 	\vspace{-0.2in}
\end{figure*}

To obtain a comparable result, we use the pre-processed dataset in \cite{UCI_data_EEG, UCI_data}. In this dataset, the original signals of 23.6 seconds were divided into 23 non-overlapping segments. Each segment contains 178 values(about 1 second). Totally, there are $5\times100\times23=11,500$ pieces of segmental signals. Figure~\ref{fft} shows five example signals from each category and their corresponding single-sided amplitude spectrum.

\begin{figure}[!ht]
\centering
	\includegraphics[scale=0.45]{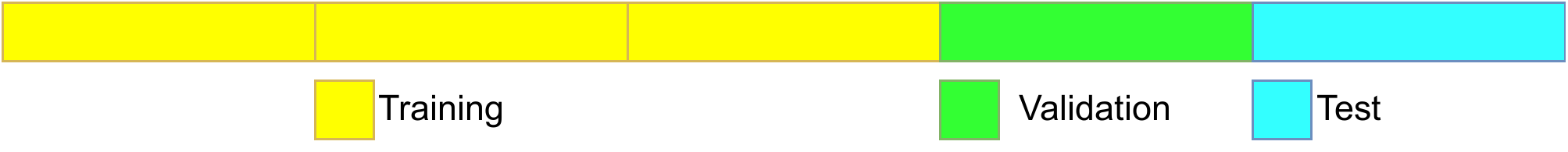}
	\vspace{-0.1in}
	\caption{Evaluation grouping detail. In the 5-fold cross validation, each group is at least once used as validation set.}
	\label{5fold}
% 	\vspace{-0.2in}
\end{figure}

To obtain the input to our neural network, we apply DFT on the original signal and only keep the first half. The right row in Figure~\ref{fft} shows the corresponding single-sided amplitude spectrum. We also follow the training steps of the 5-fold cross-validation mentioned in \cite{main_compare}. We randomly shuffle the data and divide them into five groups equally. We use one out of five as the test set, and in the other four groups, we choose one as the validation set and the other three as the training set(See Figure~\ref{5fold}). We then train our neural network with these three training sets and use the validation set to choose the number of epochs yielding the highest accuracy. After that, we retrain the neural network on the training sets and the validation set together and use the number of epochs chosen in the previous step. We then apply this final model to the test set to get the classification results. For one round of the above steps, we can get five accuracies and we keep the average of them. Then we repeat this process ten times to get the mean and standard deviation of the accuracies for each classification task. The last row in table~\ref{Comparison} shows the results obtained from our method.

\subsubsection{Result and comparison}

We can see from Table~\ref{Comparison} that our method can reach the same accuracy (99.8\%$\pm$0.13\%) on Z vs S and higher accuracy on Z vs N vs S (97.05$\pm$0.37\%) and Z vs F vs S (96.36$\pm$0.63\%) compare with the previous results. Another main concern about our method is the number of parameters and memory use. We summarize our neural network framework's parameters and memory use and the one used in \cite{main_compare} in Table~\ref{para_ram}. Our model uses 52\% fewer parameters and about 15\% more memory than the model used in \cite{main_compare}. The additional memory cost happens in the layer of taking the modulus since this layer needs to store the modulus for the use of the next layer.

\begin{table}[!ht]
\scriptsize
	\begin{center}
		\begin{tabular}{ c | c | c | c | c}
            Methods&Z vs S&S vs NS&Z vs N vs S&Z vs F vs S\\
            \hline
            \hline
            Neural Fuzzy Network\cite{Sadati_2006}&--&--&--&86.0$\pm$0.82\%\\
            Ensemble Classifier\cite{Abualsaud_ensemble_2015}&--&--&90.0$\pm$0.71\%&--\\
            Local Binary Pattern\cite{kumar_classification_2015}&--&98.3$\pm$0.24\%&--&--\\
            Multi-Domain Feature Extraction\cite{wang_automatic_2017}&--&99.0$\pm$0.15\%&--&--\\
            Exponential energy + SVM\cite{o.k._time-domain_2019}&99.5$\pm$0.20\%&--&91.7$\pm$0.65\%&89.0$\pm$0.74\%\\
            Raw data + DNN~\cite{main_compare}
            &99.0$\pm$0.29\%&98.8$\pm$0.20\%&89.2$\pm$0.73\%&89.4$\pm$0.73\%\\
            only FFT + DNN~\cite{main_compare}
            &99.8$\pm$0.13\%&99.4$\pm$0.14\%&96.3$\pm$0.45\%&95.6$\pm$0.48\%\\
            All transform + DNN\cite{main_compare}
            &99.8$\pm$0.13\%&\textbf{99.5$\pm$0.13\%}&96.5$\pm$0.44\%&95.7$\pm$0.48\%\\
            %Real+Real  &99.47\%$\pm$0.23\% &98.34\%$\pm$0.29\%&95.96\%$\pm$0.51\%&95.50\%$\pm$0.16\%\\
            Our method &\textbf{99.8$\pm$0.13\%}&98.8$\pm$0.23\%&\textbf{97.05$\pm$0.37\%}&\textbf{96.36$\pm$0.63\%}
		\end{tabular}
	\end{center} 
	\vspace{-0.2in}
	\caption{Accuracies comparison with previous works}
	\label{Comparison}
% 	\vspace{-0.1in}
\end{table}

\begin{table}[!ht]
\scriptsize
	\begin{center}
		\begin{tabular}{ c | c | r || c | c | r }
			Model in \cite{main_compare}  & \# Para & Memory  & Our method & \# Para & Memory \\ 
			\hline 
			 Input &   & 1$\times$128=0.128k & Input         &   & 128$\times$2=0.256k\\  
			       &   &   & $\mathbb{C}$ Conv & 48  & 128$\times$2$\times$8=2.048k\\ 
			       &   &   & ABS           &   & 128$\times$8=1.024k\\
			 $\mathbb{R}$ Conv  & 192  & 128$\times$32 = 4.096k  & $\mathbb{R}$ Conv  & 656  & 128$\times$16=2.048k\\
			 Pool  &   & 42$\times$32 = 1.344k  & Pool          &   & 42$\times$16=0.672k\\
			 $\mathbb{R}$ Conv  & 5152  & 42$\times$32 = 1.344k  & $\mathbb{R}$ Conv  & 2592  & 42$\times$32=1.344k\\  
			 Fc1   & 241792  & 128=0.128k  & Fc1           & 98688  & 256 = 0.256k\\
			 Fc2   & 4128  & 32=0.032k  & Fc2           & 16640  & 32 = 0.032k\\  
			 Fc3   & 128 & 3=0.003k  & Fc3           & 256  & 3 = 0.003k\\  
			\hline
			\# Total& 251392 & \textbf{7.075k} && \textbf{118880} & 8.163k
		\end{tabular}
	\end{center} 
	\vspace{-0.2in}
	\caption{Parameter and memory comparison between our method and method in \cite{main_compare}. $\mathbb{C}$ Conv: complex-valued convolutional layer, $\mathbb{R}$ Conv: real-valued convolutional layer, ABS: the layer of taking the modulus (absolute value), Pool: max pooling layer, Fc: fully connected layer.}
	\label{para_ram}
% 	\vspace{-0.1in}
\end{table}

\section{Discussion}

In this paper, we proposed a novel neural network architecture that can capture the phase information in signals by using a complex-valued convolutional layer at the very beginning. In simulations, our framework significantly improves the classification performance compared with other methods; furthermore, our method can reduce the number of parameters and improve the accuracy simultaneously in the experiments for the real-world dataset. Besides, our framework can also be applied to find proper complex-valued filters on the frequency domain without prior knowledge, which is usually a tricky task. Currently, all the input signals to our neural network are relatively short and only contain one channel. In the future, we plan to improve our method such that it can be applied to classify long-term and multi-channel EEG signals.

\section*{Availability of data and materials}

Our codes for building neural networks are based on this work \cite{cnn-from-scratch} and are available at: \url{https://github.com/David-Hang-Du/HCVNN-EEG}. The codes for simulations can also be found in our GitHub repository. 

\printbibliography

@ARTICLE{crelu,
	author={Chiheb Trabelsi and Olexa Bilaniuk and Ying Zhang and Dmitriy Serdyuk and Sandeep Subramanian and João Felipe Santos and Soroush Mehri and Negar Rostamzadeh and Yoshua Bengio and Christopher J Pal},
	journal={arXiv preprint: arXiv:1705.09792 }, 
	title={Deep Complex Networks}, 
	year={2017},
	volume={},
	number={},
	pages={},}

@article{RalphEEG,
	author = {Andrzejak, Ralph and Lehnertz, Klaus and Mormann, Florian and Rieke, Christoph and Elger, Christian},
	year = {2002},
	month = {01},
	pages = {061907},
	title = {Indications of nonlinear deterministic and finite-dimensional structures in time series of brain electrical activity: Dependence on recording region and brain state},
	volume = {64},
	journal = {Physical review. E, Statistical, nonlinear, and soft matter physics},
}

@INPROCEEDINGS{Sadati_2006,
	author={N. {Sadati} and H. R. {Mohseni} and A. {Maghsoudi}},
	booktitle={2006 IEEE International Conference on Fuzzy Systems}, 
	title={Epileptic Seizure Detection Using Neural Fuzzy Networks}, 
	year={2006},
	volume={},
	number={},
	pages={596-600}}

@article{adam,
author = {Kingma, Diederik and Ba, Jimmy},
year = {2014},
month = {12},
pages = {},
title = {Adam: A Method for Stochastic Optimization},
journal = {International Conference on Learning Representations}
}

@InProceedings{xavier,
  title = 	 {Understanding the difficulty of training deep feedforward neural networks},
  author = 	 {Xavier Glorot and Yoshua Bengio},
  booktitle = 	 {Proceedings of the Thirteenth International Conference on Artificial Intelligence and Statistics},
  pages = 	 {249--256},
  year = 	 {2010},
  volume = 	 {9},
  series = 	 {Proceedings of Machine Learning Research},
  month = 	 {05},
  publisher =    {JMLR Workshop and Conference Proceedings},
}

@article{main_compare,
title = "Classification of epileptic EEG recordings using signal transforms and convolutional neural networks",
journal = "Computers in Biology and Medicine",
volume = "109",
pages = "148 - 158",
year = "2019",
issn = "0010-4825",
author = "Rubén San-Segundo and Manuel Gil-Martín and Luis Fernando D'Haro-Enríquez and José Manuel Pardo",
}

@article{non-stationary1998,
title = {Non-stationary signal classification using the joint moments of time-frequency distributions},
journal = "Pattern Recognition",
volume = "31",
number = "11",
pages = "1635 - 1641",
year = "1998",
issn = "0031-3203",
author = {Berkant Tacer and Patrick J. Loughlin},
}

@article{ns_svm_2002,
author = {Davy, Manuel and Gretton, Arthur and Doucet, Arnaud and Rayner, Peter},
year = {2003},
month = {01},
pages = {442 - 445},
title = {Optimized support vector machines for nonstationary signal classification},
volume = {9},
journal = {Signal Processing Letters, IEEE},
}

@ARTICLE{tfa_dnn,
  author={A. T. {Tzallas} and M. G. {Tsipouras} and D. I. {Fotiadis}},
  journal={IEEE Transactions on Information Technology in Biomedicine}, 
  title={Epileptic Seizure Detection in EEGs Using Time–Frequency Analysis}, 
  year={2009},
  volume={13},
  number={5},
  pages={703-710},}

@INPROCEEDINGS{tf_svm_2014,
  author={L. {Boubchir} and S. {Al-Maadeed} and A. {Bouridane}},
  booktitle={2014 IEEE International Conference on Acoustics, Speech and Signal Processing (ICASSP)}, 
  title={On the use of time-frequency features for detecting and classifying epileptic seizure activities in non-stationary EEG signals}, 
  year={2014},
  volume={},
  number={},
  pages={5889-5893},
}

@article{QTFD_2015,
author = {Boashash, Boualem and Azemi, Ghasem and Khan, Nabeel},
year = {2015},
month = {03},
pages = {616–627},
title = {Principles of time–frequency feature extraction for change detection in non-stationary signals: Applications to newborn EEG abnormality detection},
volume = {48},
journal = {Pattern Recognition},
}

@article{tf_spatial_conv,
author = {Zhao, Dongye and Tang, Fengzhen and Si, Bailu and Feng, Xisheng},
year = {2019},
month = {03},
pages = {},
title = {Learning joint space-time-frequency features for EEG decoding on small labeled data},
volume = {114},
journal = {Neural Networks},
}

@article{GLCM_2016,
author = {Alcin, Omer and Siuly, Siuly and Bajaj, Varun and Guo, Yanhui and Sengur, Abdulkadir and Zhang, Yanchun},
year = {2016},
month = {08},
pages = {},
title = {Multi-category EEG signal classification developing Time-Frequency Texture Features based Fisher Vector encoding method},
volume = {218},
journal = {Neurocomputing},
}

@INPROCEEDINGS{motion_eeg,  author={M. R. {Islam} and M. {Ahmad}},  booktitle={2019 International Conference on Electrical, Computer and Communication Engineering (ECCE)},   title={Wavelet Analysis Based Classification of Emotion from EEG Signal},   year={2019},  volume={},  number={},  pages={1-6},}

@article{Guberman_2016_on_complex,
  author    = {Nitzan Guberman},
  title     = {On Complex Valued Convolutional Neural Networks},
  journal   = {CoRR},
  volume    = {abs/1602.09046},
  year      = {2016},
  url       = {http://arxiv.org/abs/1602.09046},
  archivePrefix = {arXiv},
  eprint    = {1602.09046},
  bibsource = {dblp computer science bibliography, https://dblp.org}
}

@article{yeung_theta_2007,
	title = {Theta phase resetting and the error-related negativity},
	volume = {44},
	issn = {0048-5772, 1469-8986},
	language = {en},
	number = {1},
	journal = {Psychophysiology},
	author = {Yeung, Nick and Bogacz, Rafal and Holroyd, Clay B. and Nieuwenhuis, Sander and Cohen, Jonathan D.},
	month = jan,
	year = {2007},
}

@article{yeung_detection_2004,
	title = {Detection of synchronized oscillations in the electroencephalogram: {An} evaluation of methods},
	volume = {41},
	issn = {0048-5772, 1469-8986},
	shorttitle = {Detection of synchronized oscillations in the electroencephalogram},
	language = {en},
	number = {6},
	journal = {Psychophysiology},
	author = {Yeung, Nick and Bogacz, Rafal and Holroyd, Clay B. and Cohen, Jonathan D.},
	month = nov,
	year = {2004},
	pages = {822--832},
}

@misc{EEG_sim_website,
	title = {Simulated {EEG} data generator},
	url = {https://data.mrc.ox.ac.uk/data-set/simulated-eeg-data-generator},
	urldate = {2021-09-12},
}

@article{makinen_auditory_2005, 
 title={Auditory event-related responses are generated independently of ongoing brain activity}, volume={24}, 
 ISSN={10538119}, 
 number={4}, 
 journal={NeuroImage}, author={M\"{a}kinen, Ville and Tiitinen, Hannu and May, Patrick}, 
 year={2005}, 
 month= feb, 
 pages={961–968} }

@article{Abualsaud_ensemble_2015, 
title={Ensemble Classifier for Epileptic Seizure Detection for Imperfect EEG Data}, 
volume={2015}, 
ISSN={2356-6140, 1537-744X}, 
journal={The Scientific World Journal}, 
author={Abualsaud, Khalid and Mahmuddin, Massudi and Saleh, Mohammad and Mohamed, Amr}, 
year={2015}, 
pages={1–15} }

@article{wang_automatic_2017, 
title={Automatic Epileptic Seizure Detection in EEG Signals Using Multi-Domain Feature Extraction and Nonlinear Analysis}, 
volume={19}, 
ISSN={1099-4300}, 
number={6},
 journal={Entropy}, 
 author={Wang, Lina and Xue, Weining and Li, Yang and Luo, Meilin and Huang, Jie and Cui, Weigang and Huang, Chao}, 
 year={2017}, 
 month=may, 
 pages={222} }

@article{kumar_classification_2015,
title = {Classification of seizure and seizure-free EEG signals using local binary patterns},
journal = {Biomedical Signal Processing and Control},
volume = {15},
pages = {33-40},
year = {2015},
issn = {1746-8094},
author = {T. Sunil Kumar and Vivek Kanhangad and Ram Bilas Pachori},
}

@inproceedings{approximate_sohn_2006,
author = {Sohn, Hansem and Lee, Wonhye and Kim, Inhye and Jeong, Jaeseung},
year = {2006},
month = {08},
pages = {1083-1086},
title = {Approximate entropy (ApEn) analysis of EEG in attention-deficit/hyperactivity disorder (ADHD) during cognitive tasks},
volume = {14},
isbn = {978-3-540-36839-7},
journal = {World Congr Proc Med Phys Biomed Eng},
}

@article{ibrahim_electroencephalography_2018,
	title = {Electroencephalography ({EEG}) signal processing for epilepsy and autism spectrum disorder diagnosis},
	volume = {38},
	issn = {02085216},
	language = {en},
	number = {1},
	journal = {Biocybernetics and Biomedical Engineering},
	author = {Ibrahim, Sutrisno and Djemal, Ridha and Alsuwailem, Abdullah},
	year = {2018},
	pages = {16--26},
}

@article{das_discrimination_2016,
	title = {Discrimination and classification of focal and non-focal {EEG} signals using entropy-based features in the {EMD}-{DWT} domain},
	volume = {29},
	issn = {17468094},
	language = {en},
	journal = {Biomedical Signal Processing and Control},
	author = {Das, Anindya Bijoy and Bhuiyan, Mohammed Imamul Hassan},
	month = aug,
	year = {2016},
	pages = {11--21},
}

@article{simons_fuzzy_2018,
	title = {Fuzzy {Entropy} {Analysis} of the {Electroencephalogram} in {Patients} with {Alzheimer}’s {Disease}: {Is} the {Method} {Superior} to {Sample} {Entropy}?},
	volume = {20},
	issn = {1099-4300},
	shorttitle = {Fuzzy {Entropy} {Analysis} of the {Electroencephalogram} in {Patients} with {Alzheimer}’s {Disease}},
	language = {en},
	number = {1},
	journal = {Entropy},
	author = {Simons, Samantha and Espino, Pedro and Abásolo, Daniel},
	month = jan,
	year = {2018},
	pages = {21},
}

@article{o.k._time-domain_2019,
	title = {Time-domain exponential energy for epileptic {EEG} signal classification},
	volume = {694},
	issn = {03043940},
	language = {en},
	journal = {Neuroscience Letters},
	author = {O.K., Fasil and R., Rajesh},
	month = feb,
	year = {2019},
	pages = {1--8},
}

@article{escalera_decoding_2010,
	title = {On the {Decoding} {Process} in {Ternary} {Error}-{Correcting} {Output} {Codes}},
	volume = {32},
	issn = {0162-8828},
	number = {1},
	journal = {IEEE Transactions on Pattern Analysis and Machine Intelligence},
	author = {Escalera, S. and Pujol, O. and Radeva, P.},
	month = jan,
	year = {2010},
	pages = {120--134},
}

@article{escalera_separability_2009,
	title = {Separability of ternary codes for sparse designs of error-correcting output codes},
	volume = {30},
	issn = {01678655},
	language = {en},
	number = {3},
	journal = {Pattern Recognition Letters},
	author = {Escalera, Sergio and Pujol, Oriol and Radeva, Petia},
	month = feb,
	year = {2009},
	pages = {285--297},
}

@article{li_detection_2018,
	title = {Detection of epileptic seizure based on entropy analysis of short-term {EEG}},
	volume = {13},
	issn = {1932-6203},
	language = {en},
	number = {3},
	journal = {PLOS ONE},
	author = {Li, Peng and Karmakar, Chandan and Yearwood, John and Venkatesh, Svetha and Palaniswami, Marimuthu and Liu, Changchun},
	editor = {Bazhenov, Maxim},
	month = mar,
	year = {2018},
	pages = {e0193691},
}

@article{richman_physiological_2000,
	title = {Physiological time-series analysis using approximate entropy and sample entropy},
	volume = {278},
	issn = {0363-6135, 1522-1539},
	language = {en},
	number = {6},
	journal = {American Journal of Physiology-Heart and Circulatory Physiology},
	author = {Richman, Joshua S. and Moorman, J. Randall},
	month = jun,
	year = {2000},
	pages = {H2039--H2049},
}

@article{li_assessing_2015,
	title = {Assessing the complexity of short-term heartbeat interval series by distribution entropy},
	volume = {53},
	issn = {0140-0118, 1741-0444},
	language = {en},
	number = {1},
	journal = {Medical \& Biological Engineering \& Computing},
	author = {Li, Peng and Liu, Chengyu and Li, Ke and Zheng, Dingchang and Liu, Changchun and Hou, Yinglong},
	month = jan,
	year = {2015},
	pages = {77--87},
}

@article{chen_measuring_2009,
	title = {Measuring complexity using {FuzzyEn}, {ApEn}, and {SampEn}},
	volume = {31},
	issn = {13504533},
	language = {en},
	number = {1},
	journal = {Medical Engineering \& Physics},
	author = {Chen, Weiting and Zhuang, Jun and Yu, Wangxin and Wang, Zhizhong},
	month = jan,
	year = {2009},
	pages = {61--68},
}

@article{shi_entropy_2017,
	title = {Entropy {Analysis} of {Short}-{Term} {Heartbeat} {Interval} {Time} {Series} during {Regular} {Walking}},
	volume = {19},
	issn = {1099-4300},
	language = {en},
	number = {10},
	journal = {Entropy},
	author = {Shi, Bo and Zhang, Yudong and Yuan, Chaochao and Wang, Shuihua and Li, Peng},
	month = oct,
	year = {2017},
	pages = {568},
}

@article{wirtinger_zur_1927,
	title = {Zur formalen {Theorie} der {Funktionen} von mehr komplexen {Ver}$\ddot{a}$nderlichen},
	volume = {97},
	issn = {0025-5831, 1432-1807},
	language = {de},
	number = {1},
	journal = {Mathematische Annalen},
	author = {Wirtinger, W.},
	month = dec,
	year = {1927},
	pages = {357--375},
}

@misc{tensorflow2015-whitepaper,
title={ {TensorFlow}: Large-Scale Machine Learning on Heterogeneous Systems},
url={https://www.tensorflow.org/},
note={Software available from tensorflow.org},
author={
    Mart\'{i}n~Abadi and
    Ashish~Agarwal and
    Paul~Barham and
    Eugene~Brevdo and
    Zhifeng~Chen and
    Craig~Citro and
    Greg~S.~Corrado and
    Andy~Davis and
    Jeffrey~Dean and
    Matthieu~Devin and
    Sanjay~Ghemawat and
    Ian~Goodfellow and
    Andrew~Harp and
    Geoffrey~Irving and
    Michael~Isard and
    Yangqing Jia and
    Rafal~Jozefowicz and
    Lukasz~Kaiser and
    Manjunath~Kudlur and
    Josh~Levenberg and
    Dandelion~Man\'{e} and
    Rajat~Monga and
    Sherry~Moore and
    Derek~Murray and
    Chris~Olah and
    Mike~Schuster and
    Jonathon~Shlens and
    Benoit~Steiner and
    Ilya~Sutskever and
    Kunal~Talwar and
    Paul~Tucker and
    Vincent~Vanhoucke and
    Vijay~Vasudevan and
    Fernanda~Vi\'{e}gas and
    Oriol~Vinyals and
    Pete~Warden and
    Martin~Wattenberg and
    Martin~Wicke and
    Yuan~Yu and
    Xiaoqiang~Zheng},
  year={2015},
}

@phdthesis{complex_sarroff_2018,
  title={Complex Neural Networks For Audio},
  author={Andy M. Sarroff},
  school= {Dartmouth College},
  year={2018}
}

@article{Gao_Gao_Chen_Liu_Zhang_2020,
 title={Deep Convolutional Neural Network-Based Epileptic Electroencephalogram (EEG) Signal Classification},
 volume={11},
 ISSN={1664-2295},
 journal={Frontiers in Neurology},
 author={Gao, Yunyuan and Gao, Bo and Chen, Qiang and Liu, Jia and Zhang, Yingchun},
 year={2020},
 pages={375} }

@article{Gao_Dang_Wang_Hong_Hou_Ma_Perc_2021,
 title={Complex networks and deep learning for EEG signal analysis},
 volume={15},
 ISSN={1871-4080, 1871-4099},
 number={3},
 journal={Cognitive Neurodynamics},
 author={Gao, Zhongke and Dang, Weidong and Wang, Xinmin and Hong, Xiaolin and Hou, Linhua and Ma, Kai and Perc, Matjaž},
 year={2021},
 month=jun,
 pages={369–388} }

@article{hassanpour_timefrequency_2004,
	title = {Time–frequency based newborn {EEG} seizure detection using low and high frequency signatures},
	volume = {25},
	issn = {0967-3334, 1361-6579},
	number = {4},
	journal = {Physiological Measurement},
	author = {Hassanpour, Hamid and Mesbah, Mostefa and Boashash, Boualem},
	month = aug,
	year = {2004},
	pages = {935--944},
}

@misc{cnn-from-scratch, 
 url={https://github.com/dreamgonfly/cnn-from-scratch}, 
 abstractNote={Building a convolutional neural network from scratch in an interactive way - cnn-from-scratch/cnn_from_scratch.ipynb at master · dreamgonfly/cnn-from-scratch}, journal={GitHub} }

@book{yamada_practical_2017,
	address = {Philadelphia, PA},
	title = {Practical {Guide} for {Clinical} {Neurophysiologic} {Testing}: {EEG}},
	isbn = {9781496383037},
	publisher = {Wolters Kluwer Health},
	author = {Yamada, Thoru and Meng, Elizabeth},
	month = oct,
	year = {2017},
}

@article{RAMGOPAL_seizure_2014,
title = {Seizure detection, seizure prediction, and closed-loop warning systems in epilepsy},
journal = {Epilepsy and Behavior},
volume = {37},
pages = {291-307},
year = {2014},
issn = {1525-5050},
author = {Sriram Ramgopal and Sigride Thome-Souza and Michele Jackson and Navah Ester Kadish and Iv\acute{a}n {S\acute{a}nchez Fern\acute{a}ndez} and Jacquelyn Klehm and William Bosl and Claus Reinsberger and Steven Schachter and Tobias Loddenkemper},
}

@phdthesis{complex_patrick_2019,
  title={Complex-valued Deep Learning with Applications to Magnetic Resonance Image Synthesis},
  author={Patrick Virtue},
  school= {University of California at Berkeley},
  year={2019}
}

@article{Scardapane_complex_2018,
author = {Scardapane, Simone and Van Vaerenbergh, Steven and Hussain, Amir and Uncini, Aurelio},
year = {2018},
month = {02},
pages = {},
title = {Complex-Valued Neural Networks With Nonparametric Activation Functions},
volume = {PP},
journal = {IEEE Transactions on Emerging Topics in Computational Intelligence},
}

@misc{bassey_2021_survey,
      title={A Survey of Complex-Valued Neural Networks}, 
      author={Joshua Bassey and Lijun Qian and Xianfang Li},
      year={2021},
      eprint={2101.12249},
      archivePrefix={arXiv},
      primaryClass={stat.ML}
}

@article{vasios_classification_2009,
	title = {Classification of {Event}-{Related} {Potentials} {Associated} with {Response} {Errors} in {Actors} and {Observers} {Based} on {Autoregressive} {Modeling}},
	volume = {3},
	issn = {18744311},
	language = {en},
	number = {1},
	journal = {The Open Medical Informatics Journal},
	author = {E. Vasios, Christos},
	month = may,
	year = {2009},
	pages = {32--43},
}

@ARTICLE{Hirose_generalization,
  author={Hirose, Akira and Yoshida, Shotaro},
  journal={IEEE Transactions on Neural Networks and Learning Systems}, 
  title={Generalization Characteristics of Complex-Valued Feedforward Neural Networks in Relation to Signal Coherence}, 
  year={2012},
  volume={23},
  number={4},
  pages={541-551},
}

@misc{UCI_data,
author = "Dua, Dheeru and Graff, Casey",
year = "2017",
title = "{UCI} Machine Learning Repository",
url = "http://archive.ics.uci.edu/ml",
institution = "University of California, Irvine, School of Information and Computer Sciences" }

@misc{UCI_data_EEG,
author = "Qiuyi Wu and Ernest Fokoue",
year = "2017",
title = "Epileptic Seizure Recognition Data Set",
url = "https://archive.ics.uci.edu/ml/datasets/Epileptic+Seizure+Recognition",
institution = "School of Mathematical Sciences
Rochester Institute of Technology" }

@article{beniczky_electroencephalography_2020,
	title = {Electroencephalography: basic biophysical and technological aspects important for clinical applications},
	volume = {22},
	issn = {1294-9361, 1950-6945},
	shorttitle = {Electroencephalography},
	number = {6},
	urldate = {2022-07-27},
	journal = {Epileptic Disorders},
	author = {Beniczky, Sándor and Schomer, Donald L.},
	month = dec,
	year = {2020},
	pages = {697--715},
}

% \bibliographystyle{plain}
% \bibliography{references.bib}

\end{document}